\documentclass[sigconf]{acmart}

\usepackage{graphicx}
\usepackage{subfigure}
\usepackage{bm}
\usepackage{enumitem}
\usepackage{amsmath}
\usepackage{amsthm}
\usepackage{multicol}
\usepackage{color}
\usepackage{multirow}
\usepackage{algorithm}
\usepackage{algorithmic}

\usepackage{amssymb}
\usepackage{wasysym}
\usepackage{xcolor}

\usepackage{balance} 

\AtBeginDocument{%
  \providecommand\BibTeX{{%
    \normalfont B\kern-0.5em{\scshape i\kern-0.25em b}\kern-0.8em\TeX}}}

\setcopyright{acmlicensed}

\copyrightyear{2024}
\acmYear{2024}
\setcopyright{acmlicensed}\acmConference[KDD '24]{Proceedings of the 30th ACM SIGKDD Conference on Knowledge Discovery and Data Mining}{August 25--29, 2024}{Barcelona, Spain}
\acmBooktitle{Proceedings of the 30th ACM SIGKDD Conference on Knowledge Discovery and Data Mining (KDD '24), August 25--29, 2024, Barcelona, Spain}
\acmDOI{10.1145/3637528.3671864}
\acmISBN{979-8-4007-0490-1/24/08}

\begin{document}

\title{Optimizing Long-tailed Link Prediction in Graph Neural Networks through Structure Representation Enhancement}



\author{Yakun Wang}
\affiliation{%
  \institution{Ant Group}
  \city{Beijing}
  \country{China}
  }
\email{feika.wyk@antgroup.com}

\author{Daixin Wang}
\affiliation{%
  \institution{Ant Group}
  \city{Beijing}
  \country{China}
  }
\email{daixin.wdx@antgroup.com}

\author{Hongrui Liu}
\affiliation{%
  \institution{Ant Group}
  \city{Beijing}
  \country{China}
  }
\email{liuhongrui.lhr@antgroup.com}

\author{Binbin Hu}
\authornote{Corresponding author.}
\affiliation{%
  \institution{Ant Group}
  \city{Hangzhou}
  \country{China}
  }
\email{bin.hbb@antfin.com}

\author{Yingcui Yan}
\affiliation{%
  \institution{Ant Group}
  \city{Shanghai}
  \country{China}
  }
\email{yanyingcui.yyc@antgroup.com}

\author{Qiyang Zhang}
\affiliation{%
  \institution{Ant Group}
  \city{Shanghai}
  \country{China}
  }
\email{buxie.zqy@antgroup.com}

\author{Zhiqiang Zhang}
\affiliation{%
  \institution{Ant Group}
  \city{Hangzhou}
  \country{China}
  }
\email{lingyao.zzq@antfin.com}



\renewcommand{\shortauthors}{Yakun Wang et al.}


\begin{abstract}
Link prediction, as a fundamental task for graph neural networks (GNNs), has boasted significant progress in varied domains.
Its success is typically influenced by the expressive power of node representation, but recent developments reveal the inferior performance of low-degree nodes owing to their sparse neighbor connections, known as the degree-based long-tailed problem.
\emph{Will the degree-based long-tailed distribution similarly constrain the efficacy of GNNs on link prediction?}
Unexpectedly, our study reveals that only a mild correlation exists between node degree and predictive accuracy, and more importantly, the number of common neighbors between node pairs exhibits a strong correlation with accuracy.
Considering node pairs with less common neighbors, \emph{i.e.,} tail node pairs, make up a substantial fraction of the dataset but achieve worse performance, we propose that link prediction also faces the long-tailed problem.
Therefore, link prediction of GNNs is greatly hindered by the tail node pairs.
After knowing the weakness of link prediction, a natural question is \emph{ how can we eliminate the negative effects of the skewed long-tailed distribution on common neighbors so as to improve the performance of link prediction?}
Towards this end, we introduce our long-tailed framework (LTLP), which is designed to enhance the performance of tail node pairs on link prediction by increasing common neighbors.
Two key modules in LTLP respectively supplement high-quality edges for tail node pairs and enforce representational alignment between head and tail node pairs within the same category, thereby improving the performance of tail node pairs.
Empirical results across five datasets confirm that our approach not only achieves SOTA performance but also greatly reduces the performance bias between the head and tail.
These findings underscore the efficacy and superiority of our framework in addressing the long-tailed problem in link prediction.
\end{abstract}




\maketitle

\section{INTRODUCTION}

Link prediction, as one of the cornerstone tasks within the realm of graph neural networks (GNNs), has been deployed in various applications such as personalized recommendation \cite{ying2018graph,he2020lightgcn,hu2018leveraging,zang2023commonsense}, drug molecule design \cite{drug_intro,drugmolecular}, and supply chain optimization \cite{yang2021financial}.
Recently, several works \cite{he2020lightgcn,kumar2020link,ying2018graph,ncnc,wang2023topological} have achieved impressive performance on link prediction, which can be approximately divided into two distinct approaches: node-centric and edge-centric. 
Node-centric methods, such as GCN-based \cite{pinsage,uniform2,gcn} and GraphSAGE-based \cite{graphsage}, focus on learning representations from individual nodes in node pairs, with the intuition that nodes with similar features or residing in similar network neighborhoods are more likely to form connections. 
Conversely, edge-centric approaches, as illustrated in works like \cite{seal,bellman-ford,buddy,neo,grail,li2024evaluating}, adeptly capitalize on the intricate structural information inherent within subgraphs (i.e. common neighbors, paths, and closed triangles), thereby garnering increased focus.
Synthesizing insights from both approaches, the performance of link prediction is typically affected by the quality of node representation and the structural information between the node pair.

\begin{figure}[htbp]
        \centering
        \subfigure[Cora Degree Analysis ]{\includegraphics[width=1.6in]{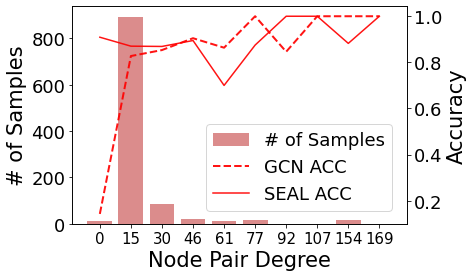} } \label{fig:cora_d}
        \subfigure[Cora CNs Analysis ]{\includegraphics[width=1.6in]{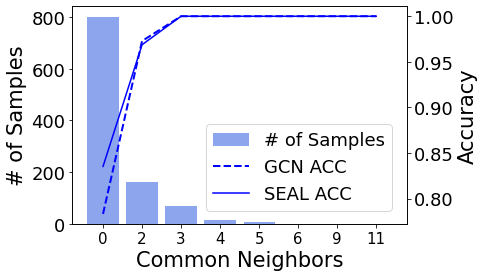}
        } \label{fig:cora_cns}
        \subfigure[CiteSeer Degree Analysis ]{\includegraphics[width=1.6in]{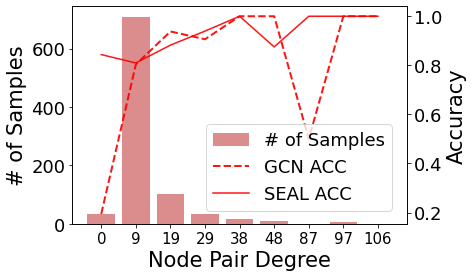}
        } \label{fig:cs_d}
        \subfigure[CiteSeer CNs Analysis ]{\includegraphics[width=1.6in]{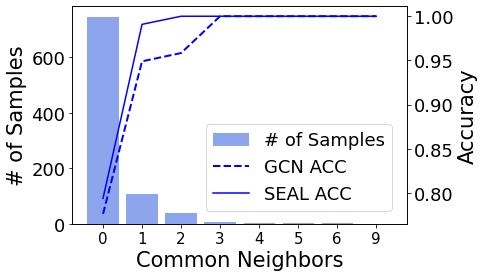}
        \label{fig:cs_cns}}
        \subfigure[Pubmed Degree Analysis ]{\includegraphics[width=1.6in]{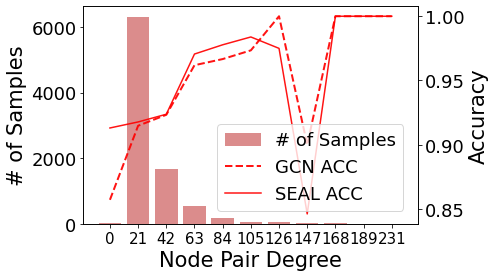}
        \label{fig:pb_d}}
        \subfigure[Pubmed CNs Analysis ]{\includegraphics[width=1.6in]{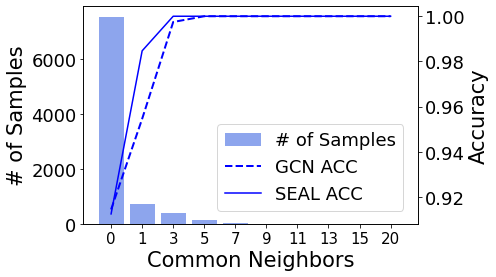}
        \label{fig:pb_cns}}
        \caption{The correlation between the link prediction accuracy and different measures, i.e. degrees and common neighbors (CNs), by using GCN and SEAL.}
        \label{fig:lt_anly}
        
\end{figure}

Despite the expressive power of these link prediction approaches, recent studies reveal that the performance of GNNs is always compromised by the long-tailed distribution of node degree, where low-degree nodes always exhibit inferior performance on classification compared with high-degree (head) nodes \cite{tailgnn,metatail2vex,coldbrew}. 
Given the importance of node representation for link prediction, the underwhelming performance of tail node pairs inspires a fundamental question: 
\emph{will the degree-based long-tailed distribution similarly constrain the efficacy of GNNs on link prediction?}
As the first contribution of this study, we present experiments to assess the relationship between node degree and accuracy in Fig. \ref{fig:lt_anly} (a)\&(c)\&(e) (more details can be seen in Sec. \ref{anly:motiva_anly}).
Surprisingly, we discover that the degree, or more specifically the degree sum of node pairs, only presents a weak or even negligible correlation with accuracy.
This unexpectedly mild correlation means that GNNs may not be as heavily influenced by the limitations in node representation when it comes to link prediction, and more importantly, it implies that there may be other underlying factors having a more pronounced impact on the link prediction capabilities of GNNs. 
This insight paves the way for novel strategies to enhance GNN models for link prediction.

Inspired by previous edge-centric methods, we propose that some specific structural information between the concerned node pairs could have an influence on the performance of link prediction.
Towards this end, we investigate the correlation between the number of common neighbors, one of structural information that is extensively leveraged on link prediction \cite{CN_m,AA,neo}, and accuracy.
Our empirical analysis, as illustrated in Fig. \ref{fig:lt_anly} (b)\&(d)\&(f), unveils a significant correlation between the number of common neighbors and accuracy for both node-centric and edge-centric methods, and more importantly, the number of common neighbors exhibits a pronounced long-tailed distribution pattern.
Therefore, link prediction also faces a similar long-tailed problem in terms of common neighbors, that node pairs with more common neighbors, \emph{i.e.,} head node pairs, achieve better results, while a majority of node pairs with less common neighbors, \emph{i.e.,} tail node pairs, achieve worse results. As a result, the tail node pairs, which make up a substantial fraction but have insufficient structural information, greatly harm the overall performance of GNN on link prediction. So one natural question arises: \emph{how can we eliminate the negative effects of the skewed long-tailed distribution on common neighbors so as to improve the performance of link prediction?}

Inspired by the strong correlation between the number of common neighbors and link prediction accuracy, we propose to increase the common neighbors of tail node pairs to improve their performance.
Intuitively, increasing common neighbors can be achieved by integrating new edges into the existing edge set, but it remains unknown how to effectively control the quality of these additional edges in case of introducing negative samples.
To address this problem, we propose our plug-in Long-Tailed Link Prediction (\textbf{LTLP}) framework, which is flexible with most GNN-based link prediction backbones.
It consists of two primary modules:
The Subgraph Enhancement Module (\textbf{SEM}) is engineered to generate high-quality relationships, harnessing a mechanism wherein the prediction score, in concert with its variance throughout the training epochs, is jointly deployed to sift through and refine a pre-defined candidate relationships. 
These additional relationships remarkably increase the common neighbors of tail node pairs and simultaneously avoid introducing negative node pairs.
As a result, it can effectively improve the performance of tail node pairs on link prediction.
The Representation Enhancement Module (\textbf{REM}) further complements the SEM module to learn a more concentrated representation for narrowing the representational disparity across all samples.
It therefore reduces the performance gap between head node pairs and tail node pairs.
Extensive experiments across various datasets and different backbones validate the effectiveness of LTLP, especially on tail node pairs.

In summary, the contributions of our work are three-fold:
\begin{itemize}[leftmargin=*]
\item We discover that link prediction faces the long-tailed problem in terms of common neighbors rather than degrees like in node classifications, where node pairs with more common neighbors achieve better results. This discovery reveals the performance bottleneck of link prediction lying in tail node pairs and inspires a new path for further improving the existing link prediction methods.

\item Owing to the strong correlation between the number of common neighbors and accuracy, we propose our plug-in long-tail link prediction framework (LTLP) to increase the common neighbors of tail node pairs for improving their performance, wherein the SEM module and REM module collectively filter out high-quality edges and achieve more concentrated representation.

\item Extensive experiments validate that LTLP not only achieves superior overall performance but also reduces the performance gap between head node pairs and tail node pairs.
The observations therefore demonstrate LTLP effectively augments the structural information by increasing common neighbors.




\end{itemize}

\section{RELATED WORK}

\subsection{Link Prediction}
Link prediction has been achieved significant success across various domains~\cite{cao2015grarep}. Among them, heuristic approaches \cite{CN_m,RA,AA,katz} assess the likelihood of connections between nodes by the variant of common neighbors and paths. Due to the expressive power of Graph Neural Networks (GNNs), GNN-based approaches \cite{he2020lightgcn,lu2011link,taskar2003link} have achieved SOTA performance. These methods are generally categorized into two categories: node-centric and edge-centric, depending on their use of subgraph structural features. Node-centric methods \cite{GCNbased,graphsage,xu2018powerful} represent nodes by gathering neighborhood information and independently combining the representations of node pairs. Despite their successes, these approaches often neglect the structural information connecting the nodes. Conversely, edge-centric methods emphasize strategies by modeling the structural information within subgraphs. For instance, SEAL \cite{seal} leverages node labeling to capture structural information, PaGNN \cite{pagnn} and NBFNet \cite{bellman-ford} automatically learn subgraph features with broadcast and aggregation operations. BUDDY \cite{buddy} represents with subgraph sketching and Neo-GNNs are identified by modeling the information of common neighbors. CFLP \cite{cflp} investigates the impact of structural information from a causal perspective and deals with counterfactual samples. NCNC \cite{ncnc} introduces structural features to guide MPNN's representation pooling and boost model performance by completing the common neighbor structure caused by graph incompleteness.
However, existing link prediction methods treat all node pairs uniformly but ignore the inherent long-tailed problem in link prediction, resulting in tail node pairs, making up a substantial fraction but achieving much poorer performance than head node pairs.

\subsection{Long-Tailed Methods in GNNs}
Given its ubiquity and innate challenges, the long-tailed problem in GNNs has also attracted significant attention. Firstly, in node-level long-tailed approaches, works like \cite{metatail2vex,tailgnn,coldbrew} build on the observation that nodes with fewer degrees, i.e. tail nodes, perform poorly, aiming to enhance the representation of tail nodes with the help of head nodes. Taking Tail-GNN as an example, this approach learns a transition matrix to bridge the gap between tail and head nodes, thereby enriching the representational capacity of tail nodes. Furthermore, SOLT-GNN \cite{longtail_graph} first states that the graph-level task also faces the long-tailed problem in terms of the graph size. Thus they employ knowledge transfer to narrow the representational gap between head and tail graphs. In link prediction, some works \cite{fairness_cons,fairlp} focus on link prediction fairness but are limited to dividing node pairs into groups based on a predefined attribute, like gender, and aim to increase performance balance among groups. By contrast, we focus on a general scenario to improve the overall performance of link prediction by enhancing the structural information of tail node pairs. 
\section{preliminary and Data analysis}

In this section, we initially formalize the task of link prediction. Next, we conduct comprehensive data analysis on three benchmark datasets, which illustrate the number of common neighbors rather than the degree is much more correlated with link prediction accuracy.
Ultimately, we uncover the long-tailed problem in link prediction, which has inspired the design of our framework.

\subsection{Link Prediction Preliminary}

Let $\mathcal{G} = \left(\mathcal{V},\mathcal{E} \right)$ be an undirected graph, with the node set $\mathcal{V} = \{v_1,v_2,v_3,\dots,v_N \}$ and the edge set $\mathcal{E}=\{(u,v)|u,v\in\mathcal{V}\}$.
Then the adjacency matrix $\mathbf{A} \in \mathbb{R}^{N \times N}$ of graph $\mathcal{G}$ is binary for simplicity where $\mathbf{A}_{uv}=1$ if there is a connection between nodes $u$ and $v$, i.e., $(u,v) \in \mathcal{E}$, otherwise, $\mathbf{A}_{uv}=0$. $\mathcal{N}_v=\{u|(u,v)\in\mathcal{E}\}$ represents the set of neighbors of node $v$, and the common neighbors between node $u$ and node $v$ can be formalized as $\mathcal{C}_{u,v}=\mathcal{N}_u \bigcap \mathcal{N}_v$.
$d_v=\sum_{u}{\mathbf{A}_{uv}}$ is denoted as the node degree of node $v$. Further, we define the node pair degree $(u,v)$ as the sum of degree of individual nodes, \emph{i.e.,} $d_{u,v}=d_u+d_v$.
Additionally, each node $v\in\mathcal{V}$  is associated with an $f$-dimension feature vector $\textbf{x}_v \in \mathbb{R}^f$. 

Graph Neural Networks (GNNs) learn node representation by leveraging an iterative aggregation function to harness the information from neighbors. 
The representation of node $v$ at the $l$-th layer can be formalized as follows:
\begin{equation}
    \bm{h}_{v}^{(l)} = \sigma (\bm{h}_{v}^{(l-1)}, \phi (\{\bm{h}_{u}^{(l-1)} | u \in \mathcal{N}_{v} \}) ),
\end{equation}
where $\bm{h}_{v}^{0}=\textbf{x}_v $, $\sigma(\cdot)$ combines the representation of node $v$ and its neighbors $\mathcal{N}_v$ of the $l$-th layer, $\phi(\cdot)$ is the aggregation function for effectively leveraging the information of neighbors.

For both node-centric and edge-centric approaches, GNN-based link prediction aims to learn a function $\mathcal{F}(u,v|\mathcal{G})$ to estimate the probability of whether a link exists between the node pair $(u, v)$. 
Formally, the prediction function $\mathcal{F}(\cdot)$ can be expressed as:
\begin{equation}
\begin{aligned}
    \mathcal{F}(u,v|\mathcal{G}) &= \textit{f} (\bm{z}_{u,v}^{(L)}), \\
    \bm{z}_{u,v}^{(L)}&=\Psi(\bm{h}_{v}^{(L)}, \bm{h}_{u}^{(L)}),
\label{eq:zuv}
\end{aligned}
\end{equation}
where the function $\textit{f}:\mathbb{R}^m\rightarrow \mathbb{R}$, which is commonly realized by the inner product or a trainable multi-layer MLP, maps an $m$-dimension output vector $\bm{z}_{u,v}^{(L)}$ to a scalar ranging from 0 to 1, representing the predicted probability of an edge existing between the two nodes.
Here, $\bm{z}_{u,v}^{(L)}\in\mathbb{R}^m$ represents an $m$-dimension hidden representation of the sample $(u,v)$, and is obtained by another function $\Psi:\mathbb{R}^m\times \mathbb{R}^m\rightarrow \mathbb{R}$, with the input $\bm{h}_u^{(L)}$ and $\bm{h}_v^{(L)}\in \mathbb{R}^{m}$ being the output of an $L$-layer GNN.
The objective of link prediction is to accurately estimate the predicted probability by $\mathcal{F}$, and therefore the objective function is

\begin{equation}
    \min_{\theta} \sum_{((u, v), y_{u,v}) \in \mathcal{O}_{train}}{\mathcal{L}(\mathcal{F}(u, v|\mathcal{G}), y_{u,v})},
    \label{eq:ori_loss}
\end{equation}
where $\mathcal{O}_{train}$ is the training set consisting of the node pair $(u,v)$ and its label $y_{u,v}$, and $\mathcal{L}(\cdot)$ is defined as the cross-entropy function following previous link prediction works \cite{seal,bellman-ford,cflp}. 
Without loss of generality, we set the label $y_{u,v}=1$ if the edge $(u,v)$ is in the edge set $\mathcal{E}$; otherwise $y_{u,v}=0$.

\begin{figure*}[t]
    \centering
    \includegraphics[width=6.4in]{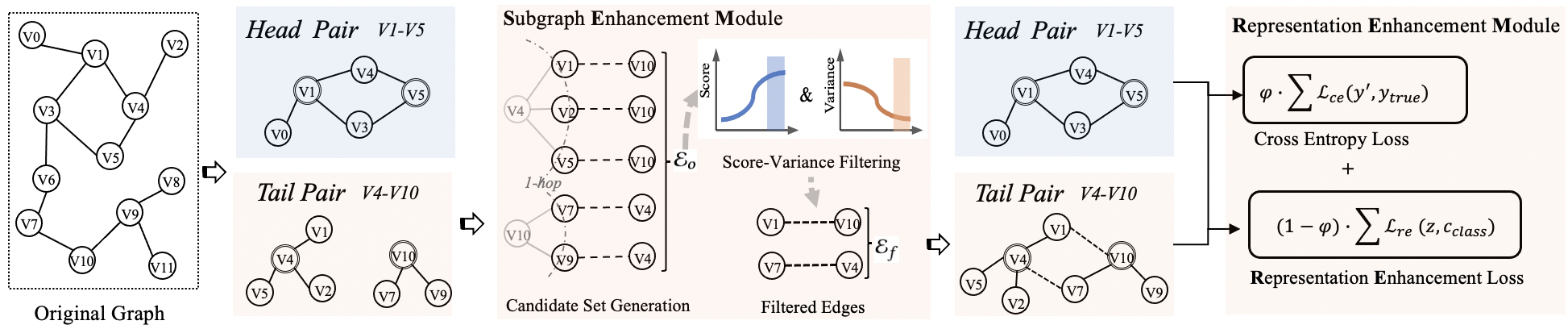}
    \caption{Overview of our proposed LTLP framework.}
    \label{fig:model_framework}
\end{figure*}

\subsection{Data Analysis}
\label{anly:motiva_anly}

As stated before, the long-tailed problem of node degree has made the representations of tail nodes underrepresented, leading to unsatisfactory performance in node classification. Therefore, we first investigate whether such degree-biased distribution will constrain the efficacy of GNNs on link prediction.
To achieve this, we conduct experiments on three benchmark datasets, \emph{i.e.,} Cora \cite{gcn}, CiteSeer \cite{gcn}, and Pubmed \cite{gcn}. In each dataset, we use the training node pairs to train two widely used link prediction models respectively of node-centric and edge-centric approaches, namely GCN and SEAL. Then we partition the test node pairs into ten uniform buckets based on the ascending order of the node pair degree $d_{u,v}$. Finally, we report the averaged link prediction accuracy and the number of samples of each bucket.
Results on Cora, Citeseer, Pubmed are respectively visualized in Fig. \ref{fig:lt_anly}(a), (c), (e). We can observe that both GCN and SEAL display few correlation between the link prediction accuracy and the degree. 

Then we continue to investigate the correlation between the
number of common neighbors, one of the structural information that
is extensively leveraged in link prediction, and the link prediction accuracy. We once again divide the test node pairs into ten buckets based on the number of their common neighbors. The link prediction accuracy and number of samples on each bucket using GCN and SEAL are reported in Fig. \ref{fig:lt_anly}(b), (d), (f), respectively for Cora, Citeseer, Pubmed.
Surprisingly, we observe a strong correlation between accuracy and the number of common neighbors on three benchmark datasets for both GCN and SEAL models. The more common neighbors they have, the higher the accuracy is. 
Furthermore, we observe that the number of common neighbors exhibits a long-tailed distribution that a large fraction of node pairs have few common neighbors. Consequently, in light of this discovery, link prediction also faces serious long-tailed problem regarding the common neighbors, which means that the structural information of tail node pairs is insufficient, thereby having a much worse performance compared to head node pairs. 
How can we eliminate the negative effects of the skewed long-tailed distribution on common neighbors so as to improve the performance of tail node pairs is critically important. 


\section{model framework}

In this section, we introduce our proposed Long-Tailed Link Prediction (LTLP) framework, which is designed to augment the structural information of tail node pairs in an effort to improve their performance on link prediction.
It is achieved by increasing common neighbors of node pairs within tail samples \footnote{We refer to the following expression of tail node pairs as tail samples in short in link prediction.}.
Specifically, as illustrated in Fig. \ref{fig:model_framework}, the LTLP framework comprises two primary modules: Structure Enhancement Module (\textbf{SEM}) and Representation Enhancement Module (\textbf{REM}).
The SEM is designed to generate high-quality relationships, harnessing a mechanism wherein the prediction score, in concert with its variance throughout the training epochs, is jointly deployed to sift through and refine pre-generated candidate relationships.
These filtered additional relationships increase the common neighbors of the tail while avoiding introducing noisy edges, thereby enhancing their representations.
To complement SEM, we propose REM to further 
narrow the representational disparity between the head and tail samples, which produces a more concentrated representation across all the samples within the same class.

\subsection{Structure Enhancement Module}
In an effort to bolster the number of common neighbors among tail samples, a straightforward strategy is to integrate new edge relationships $\mathcal{E}_{f}$ into the existing edge set $\mathcal{E}$. 
However, negative node pairs are easily introduced. 
Towards this end, we design three sub-modules, where the Candidate Set Generator module efficiently generates a candidate edge set and the subsequent Score Filtering module and Variance Filtering module collectively filter out the high-quality edge relationships. 
Such relationships increase common neighbors and are believed to improve the performance of tail samples on link prediction.

\subsubsection{Candidate Set Generator}
Considering a connected graph, the candidate edge set for the tail sample $(u,v)$ that helps for increasing common neighbors can be constructed by pairing any other nodes in the node set $\mathcal{V}$.
Therefore, for $M$ tail samples, the generation of the candidate edge set takes $(N\times M)$ time. 
In light of the potentially overwhelming number of nodes within a graph, such an exhaustive approach places a substantial demand on the computational resources, which is infeasible and can greatly burden the downstream stages.
Inspired by the observations in Sec. \ref{anly:motiva_anly}, which highlights a notable correlation between the number of common neighbors shared by a pair of nodes and the accuracy of link prediction, we advocate for a more selective approach in constructing the candidate edge set.
Specifically, we propose that the edge candidates should be restricted to those that can increase common 1-order neighbors.
Mathematically, for any link prediction sample $(u,v)$, the candidate edge set $\mathcal{E}_o$ is constructed as:
\begin{equation}
    \mathcal{E}_{o} = \bigcup_{((u,v),y_{u,v})\in\mathcal{O}} \{(u,i)|i \in \mathcal{N}_{v}\} \cup \{(v,j)|j \in \mathcal{N}_{u}\}, 
    \label{eq:edge_ori}
\end{equation}
where $\mathcal{O}$ is the dataset consisting of the node pair $(u,v)$ and its label $y_{u,v}$. Clearly, this method significantly reduces the computational overhead and enhances the efficiency of the subsequent filtering operation performed by the SVF module.
In Sec. \ref{anly:motiva_anly}, we experiment to show that only considering the common 1-order neighbors still exhibits a significant performance gain across various datasets.

\subsubsection{Score Filtering Module}
\label{subsubsec:score}
With the candidate edge set, judicious selection becomes imperative; otherwise, negative samples would be introduced to the edge set, thereby harming the performance on link prediction.
An effective approach to filter negative samples would be simply incorporating the most confidently predicted candidate edges into the existing edge set $\mathcal{E}$.
Toward this, we train a link prediction model $\mathcal{F}(u,v|\mathcal{G})$ in advance and then infer the linking probability $p_{u,v} = \mathcal{F}(u,v|\mathcal{G})$ of each candidate edge $(u,v)$ within the set $\mathcal{E}_o$.
Candidate edges that yield a predictive probability satisfying $p_{u,v}\geq \epsilon$ are selected for adding the score filtering edge set $\mathcal{E}_{s}$, where  $\epsilon$ is a pre-defined threshold.
Here, the threshold $\epsilon$ is always determined by the optimal AUC threshold of the validated set.

Despite the wide deployment of the score-based filtering strategy, we present experiments to show that it could lead to the unexpected incorporation of many false positives, thereby introducing noisy information into the graph.
Specifically, we train a link prediction model Neo-GNN on the Cora dataset \cite{gcn} following \cite{hae,vae,bellman-ford}, where we randomly sample 85\% of the data for training, 5\% for validation, and the rest 10\% for test.
Then we aim to investigate the discriminative ability of positive samples with negative samples of varying difficulty levels.
The larger the difficulty level is, the higher the probability the negative samples are incorrectly predicted by $\mathcal{F}(u,v|\mathcal{G})$.
To achieve this, we create different sets of negative samples, each corresponding to a distinct difficulty level.
For difficultly level $S=s$, we generate a set $Neg_s$ by pairing node $u$ belonging to each positive sample $(u,v)$ with $s$ randomly chosen candidate negative nodes $j$, simultaneously ensuring that $j\notin\mathcal{N}_u$.
From these candidates, we select the negative sample $(u,j)$ that yields the highest predictive score as per the model.
We record the trend of label error rate $R_{ler}=\frac{|\{(u,j)\in Neg_s|p_{u,j}\geq \epsilon\}|}{|Neg_s|}$ across training epochs.
Here, $R_{ler}$ represents the proportion of negative samples that were incorrectly scored above the threshold $\epsilon$, suggesting they are false positives. 
Clearly, from Fig. \ref{fig:data_explor} (a) we can observe that for negative samples of the highest difficulty level (green line), there are approximately 80\% samples are incorrectly predicted.
Therefore, if we were to indiscriminately accept all high-scoring predictions above the threshold, it would inevitably lead to a significant influx of noisy edges into the graph, thereby hindering the link prediction.
This trend sheds light on the limitations of the score filtering module and informs the necessity and urgency of additional filtering strategies for maintaining accuracy.

\subsubsection{Variance Filtering Module.}

\begin{figure*}[htbp]
\setlength{\abovecaptionskip}{-0.5pt}
\setlength{\belowdisplayskip}{-1cm}
        \centering
        \subfigure[Comparing of $R_{ler}$]
        {\includegraphics[width=2in]{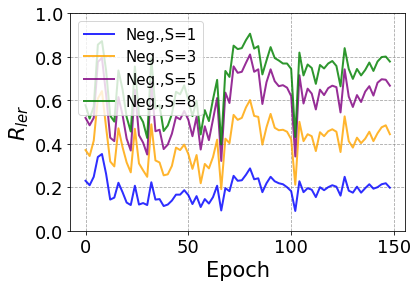}}
        \subfigure[Comparing of Variance]
        {\includegraphics[width=2in]{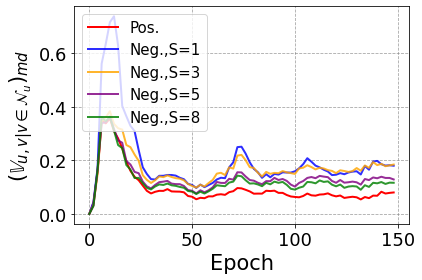}}
        \subfigure[Comparing of $R_{ler}$ after Varince filtering]
        {\includegraphics[width=2in]{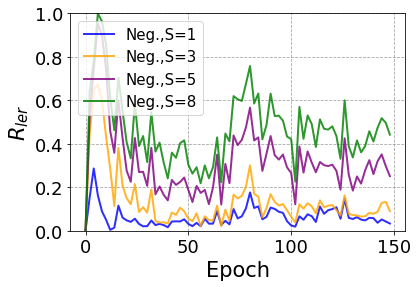}}
        \caption{Analysis of positive and negative samples on Cora dataset. $S$ denote the difficult level, $(\mathbb{V}_{u,v|v \in \mathcal{N}_{u}})_{md}$ means the median normalized variance of each epoch on set ${(u,v)|v\in\mathcal{N}_u)}$, $R_{ler}$ is the label error ratio.}
        \label{fig:data_explor}
\vspace{-5pt}
\end{figure*}

Inspired by previous works that utilize the variance of score to identify incorrectly predicted samples \cite{ding2020simplify,kendall2017uncertainties}, we hypothesize that incorporating the variance of score across epochs can also effectively filter the false positives that are incorrectly predicted by the score filtering module.
Intuitively, the variance reflects the stability of predictions during training, and therefore, the negative samples with the highest difficulty level are inclined to exhibit higher variance in their scores compared to other negative samples and the true positive samples.
In order to differentiate between positive and negative samples, we introduce the concept of normalized variance, which is the variance of the scores normalized by their mean value. 
This normalization means that negative samples with lower scores will have a proportionally higher normalized variance compared to positive samples, even if the absolute variances are similar.
Mathematically, for each candidate edge $(u,v)$ during the $t$-th epoch, the normalized variance $\mathbb{V}_{u,v}(t)$ of the score across 5 training epochs is defined as follows:
\begin{equation}
\label{eq:cal_var}
\begin{aligned}
    \mathbb{V}_{u,v}(t)&=\frac{1}{\bar{p}_{u,v}(t)}\sqrt{\frac{1}{5}\sum_{i=1}^5 \left(p_{u,v}(t-i)-\bar{p}_{u,v}(t)\right)^2},\\
    \bar{p}_{u,v}(t)&=\frac{1}{5}\sum_{i=1}^5p_{u,v}(t-i),
\end{aligned}
\end{equation}
where $p_{u,v}(i)$ represents the predictive probability of $\mathcal{F}(u,v|\mathcal{G})$ in the $i$-th epoch.
To validate the efficacy of our proposed normalized variance for filtering false negatives, we replicate the experimental setup laid out in Sec. \ref{subsubsec:score}, focusing on the separability between positive samples and negative samples of different levels of difficulty. 
We first visualized the trend of the median variance within each set of negative samples as well as the set of positive samples across epochs in Fig. \ref{fig:data_explor} (b).
We can clearly observe that the median variance is consistently lower than that of negative samples, particularly for those at the highest difficulty level.
It indicates that the normalized variance can indeed serve as a reliable indicator to separate true positives from false negatives.
Subsequently, we display the trend of label error rate $R_{ler}$ for each set of negative samples after discarding those with the 60\% lowest variance.
The results are visualized in Fig. \ref{fig:data_explor} (c). which depicts a significant improvement in the reduction of false positives when the variance-based filtering criterion is applied in conjunction with score-based filtering.
Remarkably, for the negative samples at the highest difficulty level (denoted by the green line), we observe a substantial decrease in the number of false positives, with a reduction rate approaching 30\%.
These findings validate our hypothesis that incorporating a variance-based filtering step can significantly filter the false negatives by reducing label error rates, especially for those samples where the model's predictions are highly variable and less reliable.


In light of the effectiveness of the normalized variance in identifying the false negatives, we propose to incorporate the variance filtering module into the score filtering module.
Both modules collectively filter the true positive edges from the candidate set $\mathcal{E}_o$ for increasing common neighbors.
Specifically, with the filtered candidate edge set $\mathcal{E}_f$ and the edge set $\mathcal{E}$ of the given graph $\mathcal{G}$, we expand the edge set $\mathcal{E}$ as follows:
\begin{equation}
\label{eq:E_s}
\begin{aligned}
    \mathcal{E}_{new}&=\mathcal{E} \cup \mathcal{E}_f, \\
    \mathcal{E}_f &= \{(u,v)\in \mathcal{E}_s|\mathbb{V}_{u,v}\geq \tau_{svf}\}, \\
    \mathcal{E}_s &= \{(u,v)\in \mathcal{E}_o |p_{u,v}\geq \tau\},
\end{aligned}
\end{equation}
where $p_{u,v}$ and $\mathbb{V}_{u,v}$ are obtained by the predictive result of $\mathcal{F}(u,v|\mathcal{G})$ in the final training epoch, the threshold $\tau$ for score filtering module is determined by the optimal AUC threshold on the validation set, and the threshold $\tau_{svf}$ for variance filtering module is obtained to ensure top $k$\% edges in $\mathcal{E}_s$ with the smallest variance are chosen.

\subsection{Representation Enhancement Module}
In light of the inferior performance of tail samples revealed in Sec.~\ref{anly:motiva_anly}, we additionally incorporate a representation enhancement module into our model to further enhance the tail samples for link prediction.
This module is engineered to narrow the representational disparity between head samples and tail samples, thereby producing a more concentrated representation across all the samples within the same class.

Technically, given the enhanced graph $\mathcal{G}_{new}=(\mathcal{V}, \mathcal{E}_{new})$ with the expanded edge $\mathcal{E}_{new}$ by Eq. \ref{eq:E_s}, for each training node pair $((u,v), y_{u,v}) \in \mathcal{O}_{train}$, we first get its hidden representation $\bm{z}_{u,v}$ using a GNN-based link prediction model following Eq. \ref{eq:zuv}. Note that both node-centric and edge-centric methods link prediction function can be used here for obtainin $z_{u,v}$ and our experiments demonstrate that our proposed method is flexible and effective to these backbones.


Building upon these representations, we then define a unique class center $c_{u,v}$ in the $m$-dimensional space for each class of samples—positive and negative. Specifically, this entails assigning a shared centroid for all positive samples, while simultaneously allocating a common centroid for all negative samples.
We aim to narrow the representation disparity across all the samples within the same class, thereby producing a more concentrated representation.
To achieve this, we propose the following regularization of representation enhancement loss $\mathcal{L}_{\text{re}}$ as:
\begin{equation}
    \mathcal{L}_{\text{re}} = \sum_{((u,v),y_{u,v}) \in \mathcal{O}_{train} }||\bm{z}_{u,v} - c_{u,v}||_2^2,
\end{equation}
Following \cite{seal,cflp,mebns}, we apply the cross-entropy loss as our task loss for link prediction.
Consequently, we optimize the following objective function to learn a link prediction model:

\begin{equation}
    \mathcal{L} = \varphi \ast  \mathcal{L}_{\text{ce}} + (1-\varphi) \ast \mathcal{L}_{\text{re}},
    \label{eq:final_loss}
\end{equation}
where  $\varphi$ is the hyperparameter to control the ratio between the task loss $\mathcal{L}_{\text{ce}}(\cdot)$ and the regularization term $\mathcal{L}_{\text{re}}(\cdot)$.


\begin{algorithm}[t]
    \caption{Training Pipeline of Our Framework}
    \label{alg:algorithm}
    \begin{algorithmic}[1] 
    \REQUIRE{Graph $\mathcal{G}=(\mathcal{V},
    \mathcal{E})$, Link Prediction Model: $\mathcal{F}(u,v|\mathcal{G})$}\\
    \ENSURE{ Model: $\mathcal{F}(u,v|\mathcal{G})$}
        \STATE \textbf{\textit{/** Model Pre-Training **/}}
        \STATE Update model parameters using Eq.~\ref{eq:ori_loss} 
        \STATE Memorize parameters states of the model in the last five epoch
        \STATE \textbf{\textit{/** Candidate set Generating **/}}
        \STATE Generate candidate edge set $\mathcal{E}_o$ by Eq.~\ref{eq:edge_ori} 
        \STATE Scoring pairs in $\mathcal{E}_o$ by pre-trained model and calculate variance through the last five epoch models with Eq.~\ref{eq:cal_var}
        \STATE Filtering $\mathcal{E}_f \leftarrow \mathcal{E}_0$ by scores and variance defined in Eq.~\ref{eq:E_s}
        \STATE Obtain enhanced $\mathcal{G}_{new}$ with $ (\mathcal{V}, \mathcal{E} \bigcup \mathcal{E}_f ) $
        \STATE \textbf{\textit{/** Model Continue-Training **/}}
        \FOR{epoch<T}
        \STATE Update model on enhanced $\mathcal{G}_{new}$ with Eq.~\ref{eq:final_loss}. 
        \ENDFOR
    \end{algorithmic}
\end{algorithm}

\subsection{Complexity Analysis}
As depicted in Alg.~\ref{alg:algorithm}, the overall training pipeline of LTLP includes two distinct training phases: the pre-training phase and the continued training phase, followed by a candidate edge set generator stage.
The time complexity of training one epoch for both the pre-training $\Omega_p(\mathcal{V},\mathcal{E})$ and continued training $\Omega_c(\mathcal{V},\mathcal{E})$ phase aligns with that of the baseline model $\Omega_b(\mathcal{V},\mathcal{E})$, and the time complexity of this two distinct training phases in LTLP is about $2 \times \Omega_b(\mathcal{V},\mathcal{E})$. Additionally, during the candidate set generator stage, we retain and utilize the models from the last five epochs for score inferring. 
For small datasets wherein training the model takes multiple epochs, the score inferring takes almost no time.
For large dataset wherein training only a few epochs leads to convergence, considering this stage does not involve back-propagation, the associated time complexity can be nearly expressed as $5 \times \Omega_b(\mathcal{V},\mathcal{E})\// 2$.
Consequently, the total time complexity of our framework is at most $4.5 \times \Omega_b(\mathcal{V},\mathcal{E})$ than the baseline, exhibiting a linear relationship.

\begin{table}
  \caption{The statistics of the datasets.}
  \label{tab:datasets}
  \begin{tabular}{lcccc}
    \toprule
    Dataset & Nodes & Edges & Avg. Degree & Density\\
    \midrule
     Cora & 2,708 & 5,278 & 3.9 & 0.14\% \\
     CiteSeer & 3,327 & 4,676 & 2.74 & 0.084\% \\
     Pubmed & 18,717 & 44,327 & 4.5 & 0.025\%\\
     OGB-Collab & 235,868 & 1,285,465 & 5.45 & 0.0046\%\\
     OGB-PPA & 576,289 & 30,326,273 & 52.62 & 0.018\%\\
    \bottomrule
  \end{tabular}
\end{table}

\section{experiment}

\begin{table*}
  \caption{Performance comparison (\%$\pm\sigma$) across seven baselines on five datasets. We use bold to indicate the best performance and underline for the second-best.}
  \vspace{-0.15cm}
  \label{tab:main_expr}
  \vspace{-0.15cm}
  \begin{tabular}{l|c|c|c|c|c|c|c|c}
    \toprule
    Datasets & CN & AA &  GCN & SAGE & SEAL & Neo-GNN & Tail-GNN & LTLP\\
    \midrule
     Cora & 73.45$\pm$0.00 & 73.63$\pm$0.00 & \underline{91.80}$\pm$0.54 & 90.44$\pm$0.79 & 91.14$\pm$1.24 & 89.10$\pm$0.43 & 87.95$\pm$1.08 & \textbf{93.16}$\pm$0.00 \\
     \midrule
     CiteSeer & 68.13$\pm$0.00 & 66.91$\pm$0.00 & 88.91$\pm$1.16 & 85.54$\pm$0.84 & 88.63$\pm$0.63 & \underline{90.96}$\pm$0.30 & 84.40$\pm$1.18 & \textbf{92.89}$\pm$0.16 \\
     \midrule
     Pubmed & 64.09$\pm$0.00 & 64.94$\pm$0.00 & 96.32$\pm$0.14 & 89.93$\pm$0.19 & \textbf{97.43}$\pm$0.21 & 95.73$\pm$0.32 & 94.17$\pm$0.58 & \underline{96.90}$\pm$0.23 \\
     \midrule
     OGB-Collab & 50.06$\pm$0.00 & 53.00$\pm$0.00 & 47.01$\pm$0.79 & 48.60$\pm$0.46 & 54.37$\pm$0.02 & \underline{57.52}$\pm$0.37 & 5.76$\pm$0.22 & \textbf{61.23}$\pm$1.15 \\
     \midrule
     OGB-PPA & 27.65$\pm$0.00 & 32.45$\pm$0.00 & 16.98$\pm$1.33 & 13.93$\pm$2.38 & 48.15$\pm$4.17 & \underline{49.13}$\pm$0.60 & >24h & \textbf{51.91}$\pm$2.09 \\
    \bottomrule
  \end{tabular}
\end{table*}

\subsection{Experiment Setup}

\subsubsection{Datasets} 
\label{data:data_split}
We evaluate the effectiveness of our framework on five benchmark datasets of varying sizes, including Cora, CiteSeer, Pubmed \cite{coracitepb}, OGB-Collad, and OGB-PPA \cite{ogb_data}. Detailed statistical information about these datasets is provided in Table ~\ref{tab:datasets}. Since no official training and testing sets are given in Cora, CiteSeer, and Pubmed, we follow  \cite{hae,vae,bellman-ford} to randomly split 85$\%$ of the data as the training set, 5$\%$ as the validation set, and 10$\%$ as the test set in datasets. For the OGB dataset (i.e. OGB-Collab and OGB-PPA), we follow the official data-splitting standards in \cite{ogb_data} to get the training, validation, and test data.

\subsubsection{Evaluation Metrics}
\label{data:data_eva}
In our study, we adapt the widely used metrics consistent with previous works \cite{ying2018graph,cflp,singh2021edge,wang2023neural} in link prediction. For datasets such as Cora, CiteSeer, and Pubmed, we utilize the Area Under the ROC Curve (AUC) as the evaluation criterion which aligns with the standard practice in \cite{seal,grail,mebns}. For the OGB dataset, we employ the official Hits metric (i.e. OGB-Collab: Hits@50, OGB-PPA: Hits@100) for evaluation.

Additionally, to verify the improvement on tail samples, we introduce tailored metrics that quantify the performance for both the head and tail samples, as well as gauge the disparity therein. Firstly, we divide the test set $\mathcal{O}_{test}$ into two sample groups: the head samples $\mathcal{O}_{test}^{h}$ and tail samples $\mathcal{O}_{test}^{t}$, where the head samples $\mathcal{O}_{test}^{h}$ are those samples with one or more common neighbors, while the remaining is considered as tail samples $\mathcal{O}_{test}^{t}$. 
We then assess the average accuracy on head samples as $Acc_{h}$ and tail samples as $Acc_{t}$ as commonly used in related works \cite{fairness_cons,longtail_graph}, considering the average score $Acc_{mean}=(Acc_{h} + Acc_{t})\//2$ as well as the bias $\textit{B}=Var(Acc_{h}, Acc_{t})$ as evaluation metrics. 


\subsubsection{Baselines} We benchmark our model (LTLP) against a range of prevalent link prediction approaches as well as approaches specifically tailored for long-tailed issues, listed as follows:
\begin{itemize}[leftmargin=*]
    \item Heuristic methods: We select Common Neighbours (CN) and Adamic-Adar (AA) for comparison.
    \item Node-centric approaches: We adopt two well-recognized node-centric approaches: GCN and GraphSAGE.
    \item Edge-centric approaches: SEAL and Neo-GNN are selected as exemplars of edge-centric link prediction approaches, with a strong track record of effectiveness in this domain.
    \item Long-Tailed learning: 
    For a comprehensive comparison, we adopt TailGNN, a typical method in tail-node learning, for link prediction by concatenating node representations that have been rectified by TailGNN.
\end{itemize}

Note that all baseline methods are implemented using their official codes on GitHub. LTLP is constructed using the Neo-GNN as the backbone for most experiments. We also demonstrate LTLP's flexibility with node-centric approach GCN by referring to Sec. \ref{sec:flexlibility}. More detailed implementations are in the appendix~\ref{code source}.

\subsection{Experiment Results}

\subsubsection{Results on Overall Performance}
We first report the results on overall performance in Table  \ref{tab:main_expr}. From the results, we have the following observations and analysis: 
\begin{itemize}[leftmargin=*]
    \item Across the majority of datasets, our proposed method (LTLP) consistently outstrips the baseline approaches. which underscores the superiority of our method in enhancing the overall performance of link prediction.
    \item Seal and NeoGNN always yield better results compared to other baseline methods, which demonstrates the criticality of modeling the node representations and the structure between the node pairs simultaneously in link prediction. 
    \item Heuristic methods perform poorly in most cases which demonstrates the significance to learn the linking patterns automatically from data. 
\end{itemize}

\subsubsection{Results on Head and Tail Samples}
Furthermore, we evaluate the capability of our method, LTLP, in handling long-tailed issues here. We divide the test node pairs into head and tail sample groups according to the definitions in Sec. \ref{data:data_split}. We then report the accuracy metric on these two groups separately, in a manner consistent with the methodologies presented in \cite{coldbrew,fairness_cons,longtail_graph}. The results in Table  \ref{tab:long_tail_expr} reveal that in all datasets, LTLP can achieve the best or second-best performance on head samples compared with baseline methods. More importantly, in all datasets, our method LTLP can significantly outperform all baseline methods in tail samples, which demonstrates LTLP's efficacy in amplifying the representation of tail samples.
Moreover, smaller bias in all datasets also proves that LTLP effectively narrows the performance disparity between head and tail samples, further corroborating its ability to address the challenge of the long-tailed problem in link prediction.

In summary, our method can greatly improve the overall performance of link prediction, especially on tail samples compared with state-of-the-art link prediction methods. This result aligns with our prior analysis that addressing the long-tailed problem in link prediction by structure and representation enhancement is very critical and valuable for link prediction. 


\begin{table}[htbp]
  \caption{Accuracy performance on Head and Tail Samples, as well as their Mean and Bias, with preference for Higher Values in Head, Tail, and Mean, and Lower Values in Bias. The best results are denoted in bold, and the second-best in underline.}
  \vspace{-0.15cm} 
  \label{tab:long_tail_expr}
  \vspace{-0.15cm} 
  \begin{tabular}{lc|c|c|c|c}
    \toprule
    \multicolumn{2}{c|}{Datasets} & GCN & Neo-GNN & Tail-GNN & LTLP \\
    \hline
     \multirow{4}{*}{Cora} & Head & 0.9659 & \underline{0.9739} & 0.9521 & \textbf{0.9957}  \\
     \cline{3-6}
            & Tail & 0.7838 & \underline{0.7669} & 0.7657 & \textbf{0.8585}  \\
            \cline{3-6}
            & Mean & \underline{0.8749} & 0.8704 & 0.8589 & \textbf{0.9271}  \\
            \cline{3-6}
            & Bias & \underline{0.0082} & 0.0107 & 0.0086 & \textbf{0.0047}  \\ 
    \hline
     \multirow{4}{*}{CiteSeer} & Head & \underline{0.9928} & 0.9865 & 0.9731 & \textbf{0.9939}  \\
     \cline{3-6}
            & Tail & \underline{0.7857} & 0.7385 & 0.7148 & \textbf{0.8429}  \\
            \cline{3-6}
            & Mean & \underline{0.8892} & 0.8625 & 0.8440 & \textbf{0.9184}  \\
            \cline{3-6}
            & Bias & \underline{0.0100} & 0.01538 & 0.0160 & \textbf{0.0059}  \\         
    \hline
     \multirow{4}{*}{Pubmed} & Head & 0.9807 & \textbf{0.9920} & 0.9826 & \underline{0.9892}  \\
     \cline{3-6}
            & Tail & \underline{0.9000} & 0.8924 & 0.8548 & \textbf{0.9302}  \\
            \cline{3-6}
            & Mean & 0.9404 & \underline{0.9423} & 0.9192 & \textbf{0.9607}  \\
            \cline{3-6}
            & Bias & \underline{0.0016} & 0.0024 & 0.0041 & \textbf{0.0009}  \\   

    \hline
     \multirow{4}{*}{OGB-Collab} & Head & 0.9934 & \underline{0.9975} & 0.9072 & \textbf{0.9976}  \\
     \cline{3-6}
            & Tail & \underline{0.9093} & 0.9023 & 0.8136 & \textbf{0.9124}  \\
            \cline{3-6}
            & Mean & \underline{0.9514} & 0.9499 & 0.8604 & \textbf{0.9550}  \\
            \cline{3-6}
            & Bias & \textbf{0.0017} & 0.0022 & 0.0021 & \underline{0.0018}  \\   
    \hline
     \multirow{4}{*}{OGB-PPA} & Head & 0.9856 & \textbf{0.9929} & >24h & \underline{0.9919}   \\
     \cline{3-6}
            & Tail & \underline{0.9700} & 0.9476 & >24h & \textbf{0.9770}  \\
            \cline{3-6}
            & Mean & \underline{0.9778} & 0.9702 & >24h & \textbf{0.9844}  \\
            \cline{3-6}
            & Bias & \underline{$\text{6e-5}$} & 0.0005 & >24h & $\textbf{5e-5}$  \\   
    \bottomrule
  \end{tabular}
\end{table}

\subsection{Model Analysis}

\subsubsection{Flexibility}
\label{sec:flexlibility}
As our prior analysis indicated in Sec.~\ref{anly:motiva_anly}, link prediction approaches, neither node-centric nor edge-centric, commonly exhibit the long-tailed issue in terms of common neighbors.
So we further conduct experiments based on a node-centric approach, i.e. GCN, and another edge-centric approach, i.e. NCNC, to demonstrate our framework's flexibility. Following the aforementioned experiment setting, we report both the overall performance and the accuracy on head and tail samples. As illustrated in Table  \ref{tab:GCN_ours}, our framework $LTLP_{GCN}$ and $LTLP_{NCNC}$ boost the performance of both overall and tail samples compared with GCN and NCNC. 
Consequently, our framework is effective in alleviating the long-tail problem for both node-level and edge-level models and achieving remarkable performance.

\begin{table}[htbp]
  \caption{Performance comparison based on GCN and NCNC backbone. Bold results represent the best.}
  \vspace{-0.13cm}
  \label{tab:GCN_ours}
  \vspace{-0.13cm}
  \begin{tabular}{lc|c|c|c|c}
    \toprule
    \multicolumn{2}{c|}{Datasets} & AUC & Head & Tail & Bias\\
    \hline
    \multirow{4}{*}{Cora} & $GCN$ & 0.9180 & \textbf{0.9659} & 0.7838 & 0.0082 \\
    \cline{3-6}
             & $LTLP_{GCN}$ & \textbf{0.9398} & 0.9596 & \textbf{0.8436} & \textbf{0.0033} \\
    \cline{2-6}
             & $NCNC$ & 0.9742 & \textbf{0.9786} & 0.9097 & 0.0012 \\
             \cline{3-6}
             & $LTLP_{NCNC}$ & \textbf{0.9793} & 0.9701 & \textbf{0.9317} & \textbf{0.0004} \\
    \hline
    \multirow{4}{*}{CiteSeer} & $GCN$ & 0.8891 & \textbf{0.9928} & 0.7857 & 0.0100 \\
    \cline{3-6}
             & $LTLP_{GCN}$ & \textbf{0.9124} & 0.9875 & \textbf{0.8026} & \textbf{0.0085} \\
    \cline{2-6}
             & $NCNC$ & 0.9741 & \textbf{1.0} & 0.9001 & 0.0024 \\
             \cline{3-6}
             & $LTLP_{NCNC}$ & \textbf{0.9778} & \textbf{1.0} & \textbf{0.9121} & \textbf{0.0019} \\
    \hline
    \multirow{4}{*}{Pubmed} & $GCN$ & 0.9632 & \textbf{0.9807} & 0.9000 & 0.0016 \\
    \cline{3-6}
         & $LTLP_{GCN}$ & \textbf{0.9713} & 0.9765 & \textbf{0.9144} & \textbf{0.0009} \\
    \cline{2-6}
             & $NCNC$ & 0.9913 & \textbf{0.9944} & 0.9502 & 0.0005 \\
             \cline{3-6}
             & $LTLP_{NCNC}$ & \textbf{0.9929} & 0.9928 & \textbf{0.9560} & \textbf{0.0003} \\
    \bottomrule
    \end{tabular}
\end{table}

\subsubsection{Ablation Study}
\begin{figure}[htbp]
\setlength{\abovecaptionskip}{-0.5pt}
\setlength{\belowdisplayskip}{-1cm}
        \centering
        \subfigure[Ablation studies within LTLP]{\includegraphics[width=1.6in]{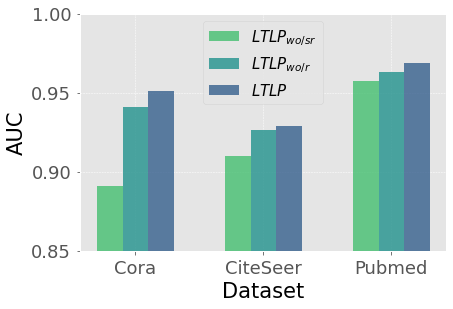}}  
        \subfigure[Ablation studies within SEM module]
        {\includegraphics[width=1.6in]{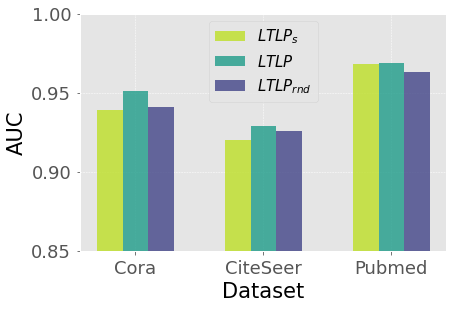}}
        \caption{Ablation studies within LTLP and SEM module.}
        \label{fig:xiaorong}
\end{figure}

Our framework consists of two primary modules: the Subgraph Enhancement Module and the Representation Enhancement Module. To validate the effectiveness of the two modules, we compare LTLP with LTLP$_{wo/r}$ and LTLP$_{wo/sr}$ on three datasets, where LTLP$_{wo/sr}$ do not contain both of the two modules, LTLP$_{wo/r}$ do not contain the Representation Enhancement Module. The result in  Fig. \ref{fig:xiaorong}, reveals that both modules contribute to the performance improvement but the Subgraph Enhancement Module can bring a more significant improvement. 

We further go into more detail on the Subgraph Enhancement Module, which consists of two submodules: the candidate edge generation and the score-variance filtering submodule. 
To evaluate the effectiveness of generating candidate common neighbors using one-hop neighbors, we use a random mechanism denoted as LTLP$_{rnd}$, where candidate common neighbors are randomly generated. We also conduct experiments by using only a score-based filtering approach, denoted as LTLP$_{s}$. The results are shown in Fig. \ref{fig:xiaorong} which demonstrate that both submodules yield better results. 

\subsubsection{Parameters Analysis}
\begin{figure}[htbp]
\setlength{\abovecaptionskip}{-0.5pt}
\setlength{\belowdisplayskip}{-1cm}
        \centering
        \subfigure[The impact of k\%]{\includegraphics[width=1.6in]{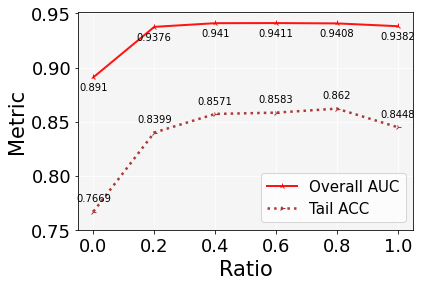}}  
        \subfigure[The impact of $\varphi$]
        {\includegraphics[width=1.6in]{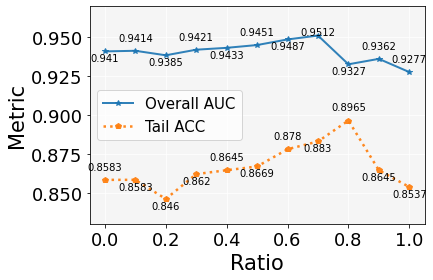}}
        \caption{Hyper-parameters analysis on Cora dataset.}
        \label{fig:paras_anly}
\end{figure}

We conduct the parameter sensitivity analysis on $K$ and $\varphi$ on Cora. The results are shown in Fig.~\ref{fig:paras_anly}. The hyper-parameter K, represents the ratio of edges maintained after the score-variance filtering mechanism. From the results, we find that the increase of $K$ before the value of $0.6$ is very essential because it can involve many common neighbors of high confidence. However, further increasing $K$ will harm the model's performance, especially on the tail samples' performance due to the involvement of false common neighbors. Additionally, the combination of representation enhancement loss and cross-entropy loss yields the best results when $\varphi=$ 0.7, and the model is robust to the value of $\varphi$.



\subsubsection{Sparsity Analysis}
To assess the efficacy of our method LTLP in sparse graph scenarios, we create sparse graph conditions by applying edge downsampling to existing graphs. Specifically, we continue to downsample the training graphs based on the settings outlined in Sec.\ref{data:data_split} on the Cora dataset.
As illustrated in Table  \ref{tab:sparse_anly} where the sampling ratio ($S$) represents the proportion of edges we randomly undersample to mimic varying levels of graph sparsity. 
Our approach, LTLP, consistently outperforms Neo-GNN in terms of the overall AUC and the accuracy of tail samples under various levels of sparsity.

\begin{table}[htbp]
  \caption{Performance comparison on the sparse graphs generated by downsampling edges on the Cora dataset, where $S$ represents the downsampling ratio, $Overall$ indicates the AUC on overall samples, and $Tail$ denotes the accuracy of tail samples.}
  \vspace{-0.13cm}
  \label{tab:sparse_anly}
  \vspace{-0.13cm}
  \begin{tabular}{lc|c|c|c|c|c}
    \toprule
    \multicolumn{2}{c|}{Methods} & $S$=0.1 & $S$=0.3 & $S$=0.5 & $S$=0.7 & $S$=0.9\\
    \hline
    \multirow{2}{*}{Neo-GNN} & $Overall$ & 0.8795 & 0.8932 & 0.8980 & 0.8982 & 0.8998\\
    \cline{3-7}
             & $Tail$ & 0.8093 & 0.8143 & 0.8055 & 0.7906 & 0.7878\\
    \hline
    \multirow{2}{*}{LTLP} & $Overall$ & 0.8904 & 0.9031 & 0.9149 & 0.9232 & 0.9271\\
    \cline{3-7}
             & $Tail$ & 0.8336 & 0.8309 & 0.8452 & 0.8571 & 0.8663\\
    \bottomrule
    \end{tabular}
\end{table}

\subsubsection{Case Study}
\begin{figure}[htbp]
        \centering
        \includegraphics[width=3.1in]{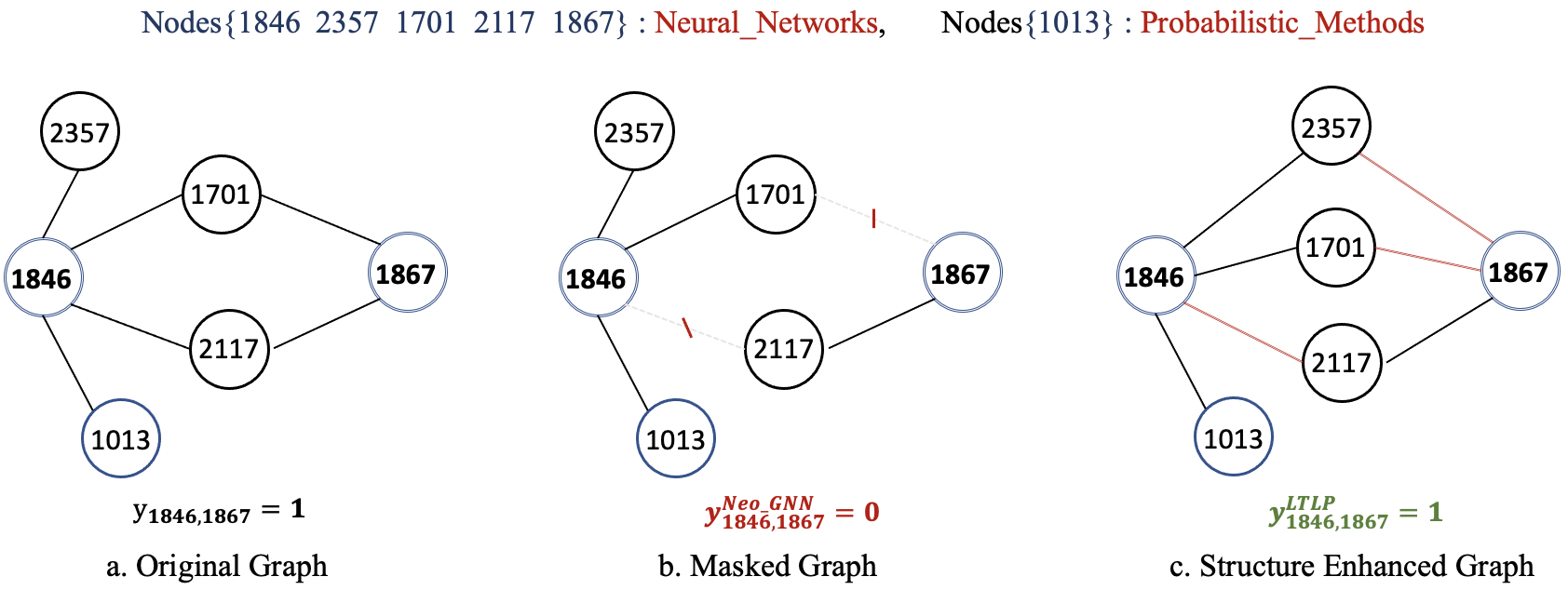}
        \caption{Case study on Cora dataset.}
        \label{fig:casestudy}
\end{figure}

To provide a more intuitive understanding of how LTLP performs and generates the common neighbors by introducing candidate edges, we conduct a case study on the Cora dataset. Consider an original graph shown in Fig.~\ref{fig:casestudy} (a), where the ground-truth label between node 1846 and node 1867 is $1$.
We simulate a scenario where a portion of the edges is masked (i.e. 50\%), thereby creating a sparser, masked graph for training the Neo-GNN model. As shown in Fig.~\ref{fig:casestudy} (b). The prediction result of Neo-GNN is wrong because of the weak structural correlation between the node pair.
We then introduce more structure information into the masked graph to form the structure enhanced graph by LTLP. As despite in Fig.~\ref{fig:casestudy} (c), we find that all of the previous common neighbors are restored by LTLP and LTLP adds one more common neighbor which has the same label with node 1846 and node 1867, resulting in a correct prediction result.  

\vfill\eject

\section{Conclusion and Future Work}

In this paper, we uncover and define the long-tailed problem in link prediction based on common neighbors. To alleviate the problem, we introduce a novel framework LTLP that enhances the representation of tail samples by supplementing common neighbors through integrating high-quality candidate edges. Moreover, we further constrain the representations of tail samples by pushing closer to the category center. Comparative experiments across multiple datasets against various baselines affirm that our method not only outperforms baselines in overall performance but also achieves more substantial improvements on tail samples, thereby substantiating the efficacy of our design. In summary, our work makes a notable contribution to the understanding of the long-tailed problem in link prediction and provides a direction for its further optimization.

Our method is relatively straightforward and effective and future work will delve deeper into the long-tailed problem in link prediction across more complex structures.


\clearpage
\bibliographystyle{ACM-Reference-Format}
\balance
\bibliography{reference}


\begin{thebibliography}{46}


\ifx \showCODEN    \undefined \def \showCODEN     #1{\unskip}     \fi
\ifx \showDOI      \undefined \def \showDOI       #1{#1}\fi
\ifx \showISBNx    \undefined \def \showISBNx     #1{\unskip}     \fi
\ifx \showISBNxiii \undefined \def \showISBNxiii  #1{\unskip}     \fi
\ifx \showISSN     \undefined \def \showISSN      #1{\unskip}     \fi
\ifx \showLCCN     \undefined \def \showLCCN      #1{\unskip}     \fi
\ifx \shownote     \undefined \def \shownote      #1{#1}          \fi
\ifx \showarticletitle \undefined \def \showarticletitle #1{#1}   \fi
\ifx \showURL      \undefined \def \showURL       {\relax}        \fi
\providecommand\bibfield[2]{#2}
\providecommand\bibinfo[2]{#2}
\providecommand\natexlab[1]{#1}
\providecommand\showeprint[2][]{arXiv:#2}

\bibitem[Adamic and Adar(2003)]%
        {AA}
\bibfield{author}{\bibinfo{person}{Lada~A Adamic} {and} \bibinfo{person}{Eytan Adar}.} \bibinfo{year}{2003}\natexlab{}.
\newblock \showarticletitle{Friends and neighbors on the web}.
\newblock \bibinfo{journal}{\emph{Social networks}}, \bibinfo{pages}{211--230}.
\newblock


\bibitem[Barab{\'a}si and Albert(1999)]%
        {CN_m}
\bibfield{author}{\bibinfo{person}{Albert-L{\'a}szl{\'o} Barab{\'a}si} {and} \bibinfo{person}{R{\'e}ka Albert}.} \bibinfo{year}{1999}\natexlab{}.
\newblock \showarticletitle{Emergence of scaling in random networks}.
\newblock \bibinfo{journal}{\emph{science}}, \bibinfo{pages}{509--512}.
\newblock


\bibitem[Cao et~al\mbox{.}(2015)]%
        {cao2015grarep}
\bibfield{author}{\bibinfo{person}{Shaosheng Cao}, \bibinfo{person}{Wei Lu}, {and} \bibinfo{person}{Qiongkai Xu}.} \bibinfo{year}{2015}\natexlab{}.
\newblock \showarticletitle{Grarep: Learning graph representations with global structural information}. In \bibinfo{booktitle}{\emph{Proceedings of the 24th ACM international on conference on information and knowledge management}}. \bibinfo{pages}{891--900}.
\newblock


\bibitem[Chamberlain et~al\mbox{.}(2023)]%
        {buddy}
\bibfield{author}{\bibinfo{person}{Benjamin~Paul Chamberlain}, \bibinfo{person}{Sergey Shirobokov}, \bibinfo{person}{Emanuele Rossi}, \bibinfo{person}{Fabrizio Frasca}, \bibinfo{person}{Thomas Markovich}, \bibinfo{person}{Nils Hammerla}, \bibinfo{person}{Michael~M Bronstein}, {and} \bibinfo{person}{Max Hansmire}.} \bibinfo{year}{2023}\natexlab{}.
\newblock \showarticletitle{Graph Neural Networks for Link Prediction with Subgraph Sketching}. In \bibinfo{booktitle}{\emph{ICLR}}.
\newblock


\bibitem[Davidson et~al\mbox{.}(2018)]%
        {hae}
\bibfield{author}{\bibinfo{person}{Tim~R Davidson}, \bibinfo{person}{Luca Falorsi}, \bibinfo{person}{Nicola De~Cao}, \bibinfo{person}{Thomas Kipf}, {and} \bibinfo{person}{Jakub~M Tomczak}.} \bibinfo{year}{2018}\natexlab{}.
\newblock \showarticletitle{Hyperspherical variational auto-encoders}.
\newblock \bibinfo{journal}{\emph{arXiv preprint arXiv:1804.00891}} (\bibinfo{year}{2018}).
\newblock


\bibitem[Ding et~al\mbox{.}(2020)]%
        {ding2020simplify}
\bibfield{author}{\bibinfo{person}{Jingtao Ding}, \bibinfo{person}{Yuhan Quan}, \bibinfo{person}{Quanming Yao}, \bibinfo{person}{Yong Li}, {and} \bibinfo{person}{Depeng Jin}.} \bibinfo{year}{2020}\natexlab{}.
\newblock \showarticletitle{Simplify and robustify negative sampling for implicit collaborative filtering}. In \bibinfo{booktitle}{\emph{NeurIPS}}. \bibinfo{pages}{1094--1105}.
\newblock


\bibitem[Hamilton et~al\mbox{.}(2017)]%
        {graphsage}
\bibfield{author}{\bibinfo{person}{Will Hamilton}, \bibinfo{person}{Zhitao Ying}, {and} \bibinfo{person}{Jure Leskovec}.} \bibinfo{year}{2017}\natexlab{}.
\newblock \showarticletitle{Inductive representation learning on large graphs}. In \bibinfo{booktitle}{\emph{NeurIPS}}.
\newblock


\bibitem[He et~al\mbox{.}(2020a)]%
        {he2020lightgcn}
\bibfield{author}{\bibinfo{person}{Xiangnan He}, \bibinfo{person}{Kuan Deng}, \bibinfo{person}{Xiang Wang}, \bibinfo{person}{Yan Li}, \bibinfo{person}{Yongdong Zhang}, {and} \bibinfo{person}{Meng Wang}.} \bibinfo{year}{2020}\natexlab{a}.
\newblock \showarticletitle{Lightgcn: Simplifying and powering graph convolution network for recommendation}. In \bibinfo{booktitle}{\emph{SIGIR}}. \bibinfo{pages}{639--648}.
\newblock


\bibitem[He et~al\mbox{.}(2020b)]%
        {uniform2}
\bibfield{author}{\bibinfo{person}{Xiangnan He}, \bibinfo{person}{Kuan Deng}, \bibinfo{person}{Xiang Wang}, \bibinfo{person}{Yan Li}, \bibinfo{person}{Yongdong Zhang}, {and} \bibinfo{person}{Meng Wang}.} \bibinfo{year}{2020}\natexlab{b}.
\newblock \showarticletitle{Lightgcn: Simplifying and powering graph convolution network for recommendation}. In \bibinfo{booktitle}{\emph{SIGIR}}. \bibinfo{pages}{639--648}.
\newblock


\bibitem[Hu et~al\mbox{.}(2018)]%
        {hu2018leveraging}
\bibfield{author}{\bibinfo{person}{Binbin Hu}, \bibinfo{person}{Chuan Shi}, \bibinfo{person}{Wayne~Xin Zhao}, {and} \bibinfo{person}{Philip~S Yu}.} \bibinfo{year}{2018}\natexlab{}.
\newblock \showarticletitle{Leveraging meta-path based context for top-n recommendation with a neural co-attention model}. In \bibinfo{booktitle}{\emph{SIGKDD}}. \bibinfo{pages}{1531--1540}.
\newblock


\bibitem[Hu et~al\mbox{.}(2020)]%
        {ogb_data}
\bibfield{author}{\bibinfo{person}{Weihua Hu}, \bibinfo{person}{Matthias Fey}, \bibinfo{person}{Marinka Zitnik}, \bibinfo{person}{Yuxiao Dong}, \bibinfo{person}{Hongyu Ren}, \bibinfo{person}{Bowen Liu}, \bibinfo{person}{Michele Catasta}, {and} \bibinfo{person}{Jure Leskovec}.} \bibinfo{year}{2020}\natexlab{}.
\newblock \showarticletitle{Open graph benchmark: Datasets for machine learning on graphs}. In \bibinfo{booktitle}{\emph{NeurIPS}}. \bibinfo{pages}{22118--22133}.
\newblock


\bibitem[Ioannidis et~al\mbox{.}(2020)]%
        {drug_intro}
\bibfield{author}{\bibinfo{person}{Vassilis~N Ioannidis}, \bibinfo{person}{Da Zheng}, {and} \bibinfo{person}{George Karypis}.} \bibinfo{year}{2020}\natexlab{}.
\newblock \showarticletitle{Few-shot link prediction via graph neural networks for covid-19 drug-repurposing}.
\newblock \bibinfo{journal}{\emph{ICML}} (\bibinfo{year}{2020}).
\newblock


\bibitem[Katz(1953)]%
        {katz}
\bibfield{author}{\bibinfo{person}{Leo Katz}.} \bibinfo{year}{1953}\natexlab{}.
\newblock \showarticletitle{A new status index derived from sociometric analysis}.
\newblock \bibinfo{journal}{\emph{Psychometrika}}, \bibinfo{pages}{39--43}.
\newblock


\bibitem[Kendall and Gal(2017)]%
        {kendall2017uncertainties}
\bibfield{author}{\bibinfo{person}{Alex Kendall} {and} \bibinfo{person}{Yarin Gal}.} \bibinfo{year}{2017}\natexlab{}.
\newblock \showarticletitle{What uncertainties do we need in bayesian deep learning for computer vision?}
\newblock \bibinfo{journal}{\emph{NeurIPS}}  \bibinfo{volume}{30} (\bibinfo{year}{2017}).
\newblock


\bibitem[Kipf and Welling(2016a)]%
        {gcn}
\bibfield{author}{\bibinfo{person}{Thomas~N Kipf} {and} \bibinfo{person}{Max Welling}.} \bibinfo{year}{2016}\natexlab{a}.
\newblock \showarticletitle{Semi-supervised classification with graph convolutional networks}. In \bibinfo{booktitle}{\emph{ICLR}}.
\newblock


\bibitem[Kipf and Welling(2016b)]%
        {vae}
\bibfield{author}{\bibinfo{person}{Thomas~N Kipf} {and} \bibinfo{person}{Max Welling}.} \bibinfo{year}{2016}\natexlab{b}.
\newblock \showarticletitle{Variational graph auto-encoders}.
\newblock \bibinfo{journal}{\emph{arXiv preprint arXiv:1611.07308}} (\bibinfo{year}{2016}).
\newblock


\bibitem[Kumar et~al\mbox{.}(2020)]%
        {kumar2020link}
\bibfield{author}{\bibinfo{person}{Ajay Kumar}, \bibinfo{person}{Shashank~Sheshar Singh}, \bibinfo{person}{Kuldeep Singh}, {and} \bibinfo{person}{Bhaskar Biswas}.} \bibinfo{year}{2020}\natexlab{}.
\newblock \showarticletitle{Link prediction techniques, applications, and performance: A survey}.
\newblock \bibinfo{journal}{\emph{Physica A: Statistical Mechanics and its Applications}} (\bibinfo{year}{2020}), \bibinfo{pages}{124289}.
\newblock


\bibitem[Li et~al\mbox{.}(2024)]%
        {li2024evaluating}
\bibfield{author}{\bibinfo{person}{Juanhui Li}, \bibinfo{person}{Harry Shomer}, \bibinfo{person}{Haitao Mao}, \bibinfo{person}{Shenglai Zeng}, \bibinfo{person}{Yao Ma}, \bibinfo{person}{Neil Shah}, \bibinfo{person}{Jiliang Tang}, {and} \bibinfo{person}{Dawei Yin}.} \bibinfo{year}{2024}\natexlab{}.
\newblock \showarticletitle{Evaluating graph neural networks for link prediction: Current pitfalls and new benchmarking}.
\newblock \bibinfo{journal}{\emph{NeurIPS}}  \bibinfo{volume}{36} (\bibinfo{year}{2024}).
\newblock


\bibitem[Li et~al\mbox{.}(2022)]%
        {fairlp}
\bibfield{author}{\bibinfo{person}{Yanying Li}, \bibinfo{person}{Xiuling Wang}, \bibinfo{person}{Yue Ning}, {and} \bibinfo{person}{Hui Wang}.} \bibinfo{year}{2022}\natexlab{}.
\newblock \showarticletitle{Fairlp: Towards fair link prediction on social network graphs}. In \bibinfo{booktitle}{\emph{AAAI}}, Vol.~\bibinfo{volume}{16}. \bibinfo{pages}{628--639}.
\newblock


\bibitem[Liu et~al\mbox{.}(2022)]%
        {longtail_graph}
\bibfield{author}{\bibinfo{person}{Zemin Liu}, \bibinfo{person}{Qiheng Mao}, \bibinfo{person}{Chenghao Liu}, \bibinfo{person}{Yuan Fang}, {and} \bibinfo{person}{Jianling Sun}.} \bibinfo{year}{2022}\natexlab{}.
\newblock \showarticletitle{On size-oriented long-tailed graph classification of graph neural networks}. In \bibinfo{booktitle}{\emph{WWW}}. \bibinfo{pages}{1506--1516}.
\newblock


\bibitem[Liu et~al\mbox{.}(2021)]%
        {tailgnn}
\bibfield{author}{\bibinfo{person}{Zemin Liu}, \bibinfo{person}{Trung-Kien Nguyen}, {and} \bibinfo{person}{Yuan Fang}.} \bibinfo{year}{2021}\natexlab{}.
\newblock \showarticletitle{Tail-gnn: Tail-node graph neural networks}. In \bibinfo{booktitle}{\emph{SIGKDD}}. \bibinfo{pages}{1109--1119}.
\newblock


\bibitem[Liu et~al\mbox{.}(2020)]%
        {metatail2vex}
\bibfield{author}{\bibinfo{person}{Zemin Liu}, \bibinfo{person}{Wentao Zhang}, \bibinfo{person}{Yuan Fang}, \bibinfo{person}{Xinming Zhang}, {and} \bibinfo{person}{Steven~CH Hoi}.} \bibinfo{year}{2020}\natexlab{}.
\newblock \showarticletitle{Towards locality-aware meta-learning of tail node embeddings on networks}. In \bibinfo{booktitle}{\emph{CIKM}}. \bibinfo{pages}{975--984}.
\newblock


\bibitem[L{\"u} and Zhou(2011)]%
        {lu2011link}
\bibfield{author}{\bibinfo{person}{Linyuan L{\"u}} {and} \bibinfo{person}{Tao Zhou}.} \bibinfo{year}{2011}\natexlab{}.
\newblock \showarticletitle{Link prediction in complex networks: A survey}.
\newblock \bibinfo{journal}{\emph{Physica A: statistical mechanics and its applications}} (\bibinfo{year}{2011}), \bibinfo{pages}{1150--1170}.
\newblock


\bibitem[Singh et~al\mbox{.}(2021)]%
        {singh2021edge}
\bibfield{author}{\bibinfo{person}{Abhay Singh}, \bibinfo{person}{Qian Huang}, \bibinfo{person}{Sijia~Linda Huang}, \bibinfo{person}{Omkar Bhalerao}, \bibinfo{person}{Horace He}, \bibinfo{person}{Ser-Nam Lim}, {and} \bibinfo{person}{Austin~R Benson}.} \bibinfo{year}{2021}\natexlab{}.
\newblock \showarticletitle{Edge proposal sets for link prediction}.
\newblock \bibinfo{journal}{\emph{arXiv preprint arXiv:2106.15810}} (\bibinfo{year}{2021}).
\newblock


\bibitem[Taskar et~al\mbox{.}(2003)]%
        {taskar2003link}
\bibfield{author}{\bibinfo{person}{Ben Taskar}, \bibinfo{person}{Ming-Fai Wong}, \bibinfo{person}{Pieter Abbeel}, {and} \bibinfo{person}{Daphne Koller}.} \bibinfo{year}{2003}\natexlab{}.
\newblock \showarticletitle{Link prediction in relational data}. In \bibinfo{booktitle}{\emph{NeurIPS}}.
\newblock


\bibitem[Teru et~al\mbox{.}(2020)]%
        {grail}
\bibfield{author}{\bibinfo{person}{Komal Teru}, \bibinfo{person}{Etienne Denis}, {and} \bibinfo{person}{Will Hamilton}.} \bibinfo{year}{2020}\natexlab{}.
\newblock \showarticletitle{Inductive relation prediction by subgraph reasoning}. In \bibinfo{booktitle}{\emph{ICML}}. \bibinfo{pages}{9448--9457}.
\newblock


\bibitem[Wang et~al\mbox{.}(2022b)]%
        {fairness_cons}
\bibfield{author}{\bibinfo{person}{Ruijia Wang}, \bibinfo{person}{Xiao Wang}, \bibinfo{person}{Chuan Shi}, {and} \bibinfo{person}{Le Song}.} \bibinfo{year}{2022}\natexlab{b}.
\newblock \showarticletitle{Uncovering the Structural Fairness in Graph Contrastive Learning}.
\newblock \bibinfo{journal}{\emph{NeurIPS}}  \bibinfo{volume}{35} (\bibinfo{year}{2022}), \bibinfo{pages}{32465--32473}.
\newblock


\bibitem[Wang et~al\mbox{.}(2023)]%
        {wang2023neural}
\bibfield{author}{\bibinfo{person}{Xiyuan Wang}, \bibinfo{person}{Haotong Yang}, {and} \bibinfo{person}{Muhan Zhang}.} \bibinfo{year}{2023}\natexlab{}.
\newblock \showarticletitle{Neural Common Neighbor with Completion for Link Prediction}.
\newblock \bibinfo{journal}{\emph{arXiv preprint arXiv:2302.00890}} (\bibinfo{year}{2023}).
\newblock


\bibitem[Wang et~al\mbox{.}(2024b)]%
        {ncnc}
\bibfield{author}{\bibinfo{person}{Xiyuan Wang}, \bibinfo{person}{Haotong Yang}, {and} \bibinfo{person}{Muhan Zhang}.} \bibinfo{year}{2024}\natexlab{b}.
\newblock \showarticletitle{Neural common neighbor with completion for link prediction}.
\newblock  (\bibinfo{year}{2024}).
\newblock


\bibitem[Wang et~al\mbox{.}(2024a)]%
        {mebns}
\bibfield{author}{\bibinfo{person}{Yakun Wang}, \bibinfo{person}{Binbin Hu}, \bibinfo{person}{Shuo Yang}, \bibinfo{person}{Meiqi Zhu}, \bibinfo{person}{Zhiqiang Zhang}, \bibinfo{person}{Qiyang Zhang}, \bibinfo{person}{Jun Zhou}, \bibinfo{person}{Guo Ye}, {and} \bibinfo{person}{Huimei He}.} \bibinfo{year}{2024}\natexlab{a}.
\newblock \showarticletitle{Not All Negatives AreWorth Attending to: Meta-Bootstrapping Negative Sampling Framework for Link Prediction}.
\newblock \bibinfo{journal}{\emph{WSDM}} (\bibinfo{year}{2024}).
\newblock


\bibitem[Wang et~al\mbox{.}(2022a)]%
        {drugmolecular}
\bibfield{author}{\bibinfo{person}{Yuyang Wang}, \bibinfo{person}{Jianren Wang}, \bibinfo{person}{Zhonglin Cao}, {and} \bibinfo{person}{Amir Barati~Farimani}.} \bibinfo{year}{2022}\natexlab{a}.
\newblock \showarticletitle{Molecular contrastive learning of representations via graph neural networks}.
\newblock \bibinfo{journal}{\emph{Nature Machine Intelligence}} (\bibinfo{year}{2022}), \bibinfo{pages}{279--287}.
\newblock


\bibitem[Wang et~al\mbox{.}(2024c)]%
        {wang2023topological}
\bibfield{author}{\bibinfo{person}{Yu Wang}, \bibinfo{person}{Tong Zhao}, \bibinfo{person}{Yuying Zhao}, \bibinfo{person}{Yunchao Liu}, \bibinfo{person}{Xueqi Cheng}, \bibinfo{person}{Neil Shah}, {and} \bibinfo{person}{Tyler Derr}.} \bibinfo{year}{2024}\natexlab{c}.
\newblock \showarticletitle{A Topological Perspective on Demystifying GNN-Based Link Prediction Performance}.
\newblock  (\bibinfo{year}{2024}).
\newblock


\bibitem[Wu et~al\mbox{.}(2020)]%
        {GCNbased}
\bibfield{author}{\bibinfo{person}{Zonghan Wu}, \bibinfo{person}{Shirui Pan}, \bibinfo{person}{Fengwen Chen}, \bibinfo{person}{Guodong Long}, \bibinfo{person}{Chengqi Zhang}, {and} \bibinfo{person}{S~Yu Philip}.} \bibinfo{year}{2020}\natexlab{}.
\newblock \showarticletitle{A comprehensive survey on graph neural networks}.
\newblock \bibinfo{journal}{\emph{IEEE transactions on neural networks and learning systems}} (\bibinfo{year}{2020}), \bibinfo{pages}{4--24}.
\newblock


\bibitem[Xu et~al\mbox{.}(2018)]%
        {xu2018powerful}
\bibfield{author}{\bibinfo{person}{Keyulu Xu}, \bibinfo{person}{Weihua Hu}, \bibinfo{person}{Jure Leskovec}, {and} \bibinfo{person}{Stefanie Jegelka}.} \bibinfo{year}{2018}\natexlab{}.
\newblock \showarticletitle{How powerful are graph neural networks?}. In \bibinfo{booktitle}{\emph{ICLR}}.
\newblock


\bibitem[Yang et~al\mbox{.}(2021a)]%
        {pagnn}
\bibfield{author}{\bibinfo{person}{Shuo Yang}, \bibinfo{person}{Binbin Hu}, \bibinfo{person}{Zhiqiang Zhang}, \bibinfo{person}{Wang Sun}, \bibinfo{person}{Yang Wang}, \bibinfo{person}{Jun Zhou}, \bibinfo{person}{Hongyu Shan}, \bibinfo{person}{Yuetian Cao}, \bibinfo{person}{Borui Ye}, \bibinfo{person}{Yanming Fang}, {et~al\mbox{.}}} \bibinfo{year}{2021}\natexlab{a}.
\newblock \showarticletitle{Inductive Link Prediction with Interactive Structure Learning on Attributed Graph}. In \bibinfo{booktitle}{\emph{ECML PKDD}}. \bibinfo{pages}{383--398}.
\newblock


\bibitem[Yang et~al\mbox{.}(2021b)]%
        {yang2021financial}
\bibfield{author}{\bibinfo{person}{Shuo Yang}, \bibinfo{person}{Zhiqiang Zhang}, \bibinfo{person}{Jun Zhou}, \bibinfo{person}{Yang Wang}, \bibinfo{person}{Wang Sun}, \bibinfo{person}{Xingyu Zhong}, \bibinfo{person}{Yanming Fang}, \bibinfo{person}{Quan Yu}, {and} \bibinfo{person}{Yuan Qi}.} \bibinfo{year}{2021}\natexlab{b}.
\newblock \showarticletitle{Financial risk analysis for SMEs with graph-based supply chain mining}. In \bibinfo{booktitle}{\emph{IJCAI}}. \bibinfo{pages}{4661--4667}.
\newblock


\bibitem[Yang et~al\mbox{.}(2016)]%
        {coracitepb}
\bibfield{author}{\bibinfo{person}{Zhilin Yang}, \bibinfo{person}{William Cohen}, {and} \bibinfo{person}{Ruslan Salakhudinov}.} \bibinfo{year}{2016}\natexlab{}.
\newblock \showarticletitle{Revisiting semi-supervised learning with graph embeddings}. In \bibinfo{booktitle}{\emph{ICML}}. \bibinfo{pages}{40--48}.
\newblock


\bibitem[Ying et~al\mbox{.}(2018a)]%
        {ying2018graph}
\bibfield{author}{\bibinfo{person}{Rex Ying}, \bibinfo{person}{Ruining He}, \bibinfo{person}{Kaifeng Chen}, \bibinfo{person}{Pong Eksombatchai}, \bibinfo{person}{William~L Hamilton}, {and} \bibinfo{person}{Jure Leskovec}.} \bibinfo{year}{2018}\natexlab{a}.
\newblock \showarticletitle{Graph convolutional neural networks for web-scale recommender systems}. In \bibinfo{booktitle}{\emph{SIGKDD}}. \bibinfo{pages}{974--983}.
\newblock


\bibitem[Ying et~al\mbox{.}(2018b)]%
        {pinsage}
\bibfield{author}{\bibinfo{person}{Rex Ying}, \bibinfo{person}{Ruining He}, \bibinfo{person}{Kaifeng Chen}, \bibinfo{person}{Pong Eksombatchai}, \bibinfo{person}{William~L Hamilton}, {and} \bibinfo{person}{Jure Leskovec}.} \bibinfo{year}{2018}\natexlab{b}.
\newblock \showarticletitle{Graph convolutional neural networks for web-scale recommender systems}. In \bibinfo{booktitle}{\emph{SIGKDD}}. \bibinfo{pages}{974--983}.
\newblock


\bibitem[Yun et~al\mbox{.}(2021)]%
        {neo}
\bibfield{author}{\bibinfo{person}{Seongjun Yun}, \bibinfo{person}{Seoyoon Kim}, \bibinfo{person}{Junhyun Lee}, \bibinfo{person}{Jaewoo Kang}, {and} \bibinfo{person}{Hyunwoo~J Kim}.} \bibinfo{year}{2021}\natexlab{}.
\newblock \showarticletitle{Neo-gnns: Neighborhood overlap-aware graph neural networks for link prediction}.
\newblock \bibinfo{journal}{\emph{NeurIPS}}, \bibinfo{pages}{13683--13694}.
\newblock


\bibitem[Zang et~al\mbox{.}(2023)]%
        {zang2023commonsense}
\bibfield{author}{\bibinfo{person}{Xiaoling Zang}, \bibinfo{person}{Binbin Hu}, \bibinfo{person}{Jun Chu}, \bibinfo{person}{Zhiqiang Zhang}, \bibinfo{person}{Guannan Zhang}, \bibinfo{person}{Jun Zhou}, {and} \bibinfo{person}{Wenliang Zhong}.} \bibinfo{year}{2023}\natexlab{}.
\newblock \showarticletitle{Commonsense Knowledge Graph towards Super APP and Its Applications in Alipay}. In \bibinfo{booktitle}{\emph{SIGKDD}}. \bibinfo{pages}{5509--5519}.
\newblock


\bibitem[Zhang and Chen(2018)]%
        {seal}
\bibfield{author}{\bibinfo{person}{Muhan Zhang} {and} \bibinfo{person}{Yixin Chen}.} \bibinfo{year}{2018}\natexlab{}.
\newblock \showarticletitle{Link prediction based on graph neural networks}. In \bibinfo{booktitle}{\emph{NeurIPS}}.
\newblock


\bibitem[Zhao et~al\mbox{.}(2022)]%
        {cflp}
\bibfield{author}{\bibinfo{person}{Tong Zhao}, \bibinfo{person}{Gang Liu}, \bibinfo{person}{Daheng Wang}, \bibinfo{person}{Wenhao Yu}, {and} \bibinfo{person}{Meng Jiang}.} \bibinfo{year}{2022}\natexlab{}.
\newblock \showarticletitle{Learning from counterfactual links for link prediction}. In \bibinfo{booktitle}{\emph{ICML}}. \bibinfo{pages}{26911--26926}.
\newblock


\bibitem[Zheng et~al\mbox{.}(2021)]%
        {coldbrew}
\bibfield{author}{\bibinfo{person}{Wenqing Zheng}, \bibinfo{person}{Edward~W Huang}, \bibinfo{person}{Nikhil Rao}, \bibinfo{person}{Sumeet Katariya}, \bibinfo{person}{Zhangyang Wang}, {and} \bibinfo{person}{Karthik Subbian}.} \bibinfo{year}{2021}\natexlab{}.
\newblock \showarticletitle{Cold brew: Distilling graph node representations with incomplete or missing neighborhoods}.
\newblock \bibinfo{journal}{\emph{arXiv preprint arXiv:2111.04840}} (\bibinfo{year}{2021}).
\newblock


\bibitem[Zhou et~al\mbox{.}(2009)]%
        {RA}
\bibfield{author}{\bibinfo{person}{Tao Zhou}, \bibinfo{person}{Linyuan L{\"u}}, {and} \bibinfo{person}{Yi-Cheng Zhang}.} \bibinfo{year}{2009}\natexlab{}.
\newblock \showarticletitle{Predicting missing links via local information}.
\newblock \bibinfo{journal}{\emph{The European Physical Journal B}}, \bibinfo{pages}{623--630}.
\newblock


\bibitem[Zhu et~al\mbox{.}(2021)]%
        {bellman-ford}
\bibfield{author}{\bibinfo{person}{Zhaocheng Zhu}, \bibinfo{person}{Zuobai Zhang}, \bibinfo{person}{Louis-Pascal Xhonneux}, {and} \bibinfo{person}{Jian Tang}.} \bibinfo{year}{2021}\natexlab{}.
\newblock \showarticletitle{Neural bellman-ford networks: A general graph neural network framework for link prediction}. In \bibinfo{booktitle}{\emph{NeurIPS}}. \bibinfo{pages}{29476--29490}.
\newblock


\end{thebibliography}

\clearpage
\appendix
\section{Appendix}

\subsection{LTLP Configuration and Baseline Settings}
\label{code source}
Our LTLP framework is based on the Neo-GNN architecture, which includes three layers of GCN and a hidden dimension of 256. More detailed configurations are in the following Neo-GNN code.
In our experiments, the baseline models primarily rely on open-source frameworks and github. The reproducible code for each baseline is available as below:

(1) CN/AA: https://github.com/facebookresearch/SEAl\_OGB/ \\
blob/main/utils.py 

(2) GCN/SAGE: https://github.com/pyg-team/pytorch\_geometric \\ /blob/master/examples/link\_pred.py 

(3) SEAL: https://github.com/facebookresearch/SEAL\_OGB  

(4) Neo-GNN: https://github.com/seongjunyun/Neo-GNNs  

(5) Tail-GNN: https://github.com/shuaiOKshuai/Tail-GNN  

(6) NCNC: https://github.com/GraphPKU/NeuralCommonNeighbor

\begin{table}[htbp]
    \centering
    \caption{Parameters setting in LTLP across datastes. $epoch_1$ means iteration epochs in the pretraining stage and $epoch_2$ is in continued training stage.}
    \begin{tabular}{l|c|c|c|c|c}
        \toprule
        Dataset & K & $\varphi$ & $epoch_1$ & $epoch_2$ & batchsize \\
        \hline
        Cora & 0.6 & 0.7 & 120 & 50 & 1024 \\
        \hline
        CiteSeer & 0.4 & 0.6 & 120 & 70 & 1024 \\ 
        \hline
        Pubmed & 0.8 & 0.2 & 60 & 60 & 1024 \\ 
        \hline
        OGB-Collab & 0.6 & 0.6 & 100 & 30 & 1024 \\
        \hline
        OGB-PPA & 0.4 & 0.2 & 100 & 50 & 1024 \\
        \bottomrule
    \end{tabular}
    \label{tab:parameter_setting}
\end{table}

\begin{figure}[htbp]
        \centering
        \subfigure[CNS distribution on Cora]{\includegraphics[width=1.6in]{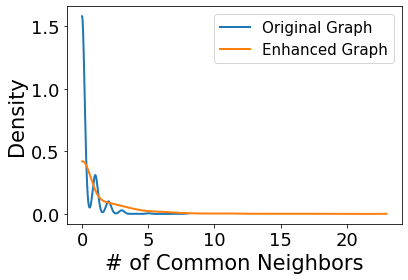} } \label{fig:cora_dis}
        \subfigure[CNS distribution on CiteSeer]{\includegraphics[width=1.6in]{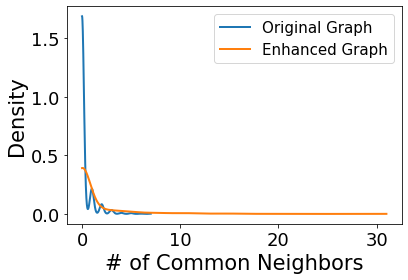}
        } \label{fig:cs_dis}
        \subfigure[CNS distribution on Pubmed]{\includegraphics[width=1.6in]{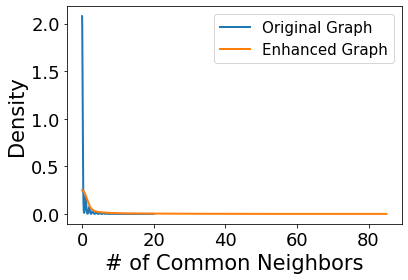}
        } \label{fig:pb_dis}
        \caption{The distribution of common neighbors (CNS) in the original graph and enhanced graph.}
        \label{fig:append_visual}
\end{figure}

\subsection{Parameters and Environments}
\label{apd:hyperparameters}
The optimal hyperparameters are determined through grid search on the validation set, which mainly includes K and $\varphi$. K controls the ratio of variance-based sample filtering, and $\varphi$ regulates the emphasis placed on the two loss functions. Additionally, there are other training parameters, such as batch size, and epochs for both the pre-training and continued training phases of LTLP, with specific settings for each dataset detailed in Table ~\ref{tab:parameter_setting}.
As for the training environment, all experiments in this paper are performed with a Tesla V100 GPU(32GB).



\subsection{Visualization on Structure Enhancement}
Building upon the analysis of the correlation between the accuracy of link prediction and common neighbors. Consequently, we proceed to examine the distribution shift of common neighbors before and after structure enhancement. We plot the distribution of common neighbors in both the original graph and the enhanced graph by edge addition. 
The findings shown in Fig. \ref{fig:append_visual} indicate that by augmenting common neighbors through the structure enhancement module, a significant number of tail samples are transformed into head samples, which in turn introduces the structural information within the subgraph and improves the precision.

\subsection{Data Analysis on OGB Datasets}
\label{append_ogb_anly}
\begin{figure}[htbp]
        \centering
        \subfigure[OGB-Collab Degree Analysis ]{\includegraphics[width=1.6in]{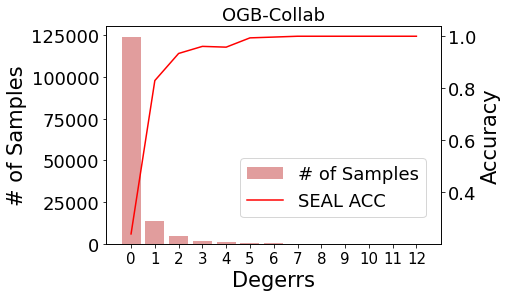} } \label{fig:collab_d}
        \subfigure[OGB-Collab CNs Analysis ]{\includegraphics[width=1.6in]{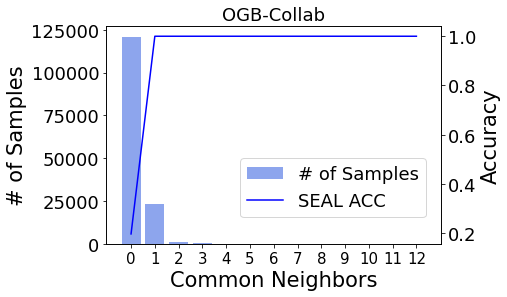}
        } \label{fig:collab_cns}
        \subfigure[OGB-PPA Degree Analysis ]{\includegraphics[width=1.6in]{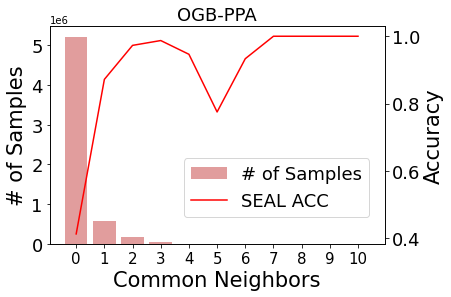}
        } \label{fig:ppa_d}
        \subfigure[OGB-PPA CNs Analysis ]{\includegraphics[width=1.6in]{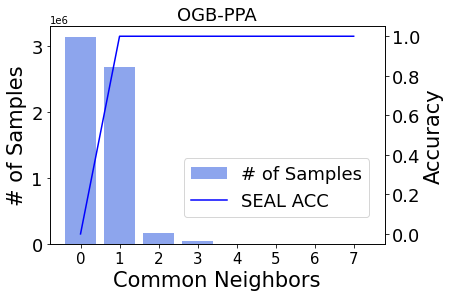}
        \label{fig:ppa_cns}}
        \caption{The correlation between the link prediction accuracy and different measures, i.e. degrees and common neighbors (CNs), by using SEAL.}
        \label{fig:append_lt_anly}
\end{figure}

We also conduct data analysis on the large-scale OGB datasets (i.e. OGB-collab, OGB-PPA), employing the same setup as described in Sec. \ref{anly:motiva_anly} and using the classic model SEAl as the baseline. As illustrated in Fig. \ref{fig:append_lt_anly}, the results align with those presented in Sec. \ref{anly:motiva_anly}, in terms of both the distribution of samples and the correlation with accuracy. This analysis further substantiates that the definition of long-tailed link prediction is associated with common neighbors rather than degrees.

\end{document}